# How to Navigate Uncertainty About AI Consciousness

**Dr Tom McClelland**[1]

**Abstract.**
Given deep uncertainty about the possibility of artificial consciousness, it is unclear how we should treat potentially sentient AI. On the one hand, we could assume insentience but risk doing terrible harms to entities that deserve moral standing. On the other hand, we could assume sentience and instead risk wasting resources on insentient machines. The intractability of questions around AI consciousness mean that this dilemma is hard to escape. I suggest a way out of that shifts from intractable questions of AI consciousness to tractable questions of AI valence. Specifically, we can assess whether an AI has states that *would* constitute valenced experiences *if* it were conscious. I show how this is sufficient to ground a responsible approach to the development of potentially conscious AI.

## 1 THE MORAL QUANDARY

The possibility of artificial consciousness (AC) leaves us in a deep moral quandary. If an AI has consciousness (or perhaps consciousness of the right kind) then it plausibly deserves our moral consideration. But if that AI lacks consciousness (or consciousness of the right kind) then it would plausibly lack moral standing. Put another way, being conscious is bound up with being a moral patient. Consciousness, or some version of it, could well mark the boundary of what Singer calls 'the moral circle': the set of all entities whose interests we are morally obliged to recognise.[2] If this credible ethical outlook is right, then questions around artificial consciousness become issues of great moral import.

Views that link moral patienthood to consciousness come in two main versions.[3] The first version says that consciousness is both *necessary* and *sufficient* for moral patienthood. If an entity has conscious experiences of any kind, then it warrants our moral consideration. The fact that there is *something it's like to be* that entity is enough for it to have moral standing. Conversely, if there is *nothing* it's like to be that entity then it not a suitable target for our moral consideration.
The second version says that consciousness *of the right kind* is necessary and sufficient for consciousness. Specifically, what matters ethically is *valenced* consciousness i.e. conscious experiences that are positive or negative to be in for the subject that undergoes them. This includes experiences of pain, pleasure, emotion and any other experience that feels good or bad for the subject. A being with valenced consciousness is *sentient*. According to 'sentientism', an entity is a moral patient if and only if it is sentient. What matters morally is the capacity for positive experiences and the capacity to suffer.

Like the first view, sentientism says that if an entity lacks consciousness then it cannot qualify as a moral patient. But unlike the first view, sentientism says that consciousness *as such* is not enough for entry into the moral circle. Consider an entity with a completely neutral phenomenology devoid of positive or negative feelings. Chalmers calls such beings '*Vulcans'*.[4] Sentientism claims that a subject like this would not qualify as a moral patient. After all, if such a subject does not experience anything as positive or negative then it is hard to see how our actions could affect it for good or ill. On this view, the Vulcan has no real interests that ought to feature in our moral deliberations and so has no moral standing.

I will assume the sentientist view that sentience is both necessary and sufficient for moral patienthood. This is not because sentientism is uncontroversial. Various authors argue that entities that are conscious but not sentient still deserve our moral consideration.[5] And other authors argue that moral patienthood has little or nothing to do with consciousness.[6] Nevertheless, sentientism is a credible view that has attracted many supporters and plays a central role in discussions of moral circle expansion.[7] My aim is to explore the challenge presented to sentientism by AI.

According to sentientism, an AI is a moral patient if and only if it has valenced experiences. Determining whether an AI is sentient, or at least how likely it is to be sentient, is then imperative to knowing how it ought to be treated. If it is sentient, or has a sufficiently high likelihood of sentience, there is a clear obligation to protect its welfare.

---

[1] Department of History and Philosophy or Science, University of Cambridge
[2] Singer, 2011.
[3] Roelofs, 2023.
[4] Chalmers, 2022.
[5] Chalmers, 2022.
[6] Kammerer, 2022.
[7] Sebo, 2025.

And if it is not sentient then it does not really have any welfare to consider and is not a moral patient.

But now we run into difficulties. Sentience requires consciousness and we face considerable uncertainty regarding the prospects of artificial consciousness.[8] This uncertainty is underwritten by deep disagreement about how to test for artificial consciousness: different theories point to different markers of consciousness and serious questions remain about the legitimacy of using those markers in an AI context. The uncertainty is also underwritten by deep disagreement about whether artificial consciousness is even possible: biological naturalists regard consciousness as a biological phenomenon so are sceptical of AC while computational functionalists regard consciousness as substrate-independent so regard AC as a live possibility. But such uncertainty about whether an AI is conscious then entails uncertainty about whether it is a sentient moral patient.

Not knowing whether a given AI is sentient leaves us with a moral dilemma regarding its treatment. On the one hand, we could act on the assumption that the AI is not sentient. If we are right, then no harm is done. But if we are wrong, we could be guilty of perpetrating great harms against the AI. Our moral decision-making would be completely insensitive to its well-being. Scaling this up over countless AIs, we could be collectively responsible for suffering on an industrial scale. On the other hand, we could act on the assumption that AI is sentient. This avoids the aforementioned risk of moral catastrophe but comes with its own moral costs. If the AI is indeed sentient, treating it as sentient would be the right thing. But if it is *not* sentient then our efforts to protect its interests would be wasted. Any resources we spend on this policy could instead have been spent on protecting genuine sentient beings. And any constraints placed on progress in the development of advanced AI could deprive countless moral patients from reaping the benefits of that progress.

My paper will proceed as follows: in Section 2 I will outline how advocates of the Precautionary Principle respond to this dilemma and offer some objections to that response. In Section 3 I will consider an alternative approach that I call the Avoidance Strategy and show how this too fails. Both approaches attempt to confront uncertainty around attributions of AC but end up themselves being compromised by issues of deep uncertainty. In Section 4 I propose a shift from questions of consciousness to questions of valence. I then show what this shift would mean for the Precautionary Principle and Avoidance Strategy in Sections 5 and 6 respectively. A revised version of the Avoidance Strategy emerges as a promising way forward. In Section 7 I give an overview of research into AI valence and the challenges it faces before concluding in Section 8.

[8] McClelland, 2025.

## 2 THE PRECAUTIONARY PRINCIPLE AND ITS DISCONTENTS

Faced with a choice between two risky options, a sensible way forward is to weigh up the seriousness of each risk. In the dilemma described above, there is a risk associated with treating AI as sentient and a risk associated with not treating it as sentient. But these two risks are not equal. The scenario where we dedicate resources to insentient machines is bad, but the scenario where we fail to protect the welfare of sentient machines is catastrophic. A case can thus be made for erring on the side of caution. This thought is encoded in the Precautionary Principle which underwrites a great deal of work on AI welfare. Birch makes a strong case for taking this approach to various ‘edge cases’ of sentience including foetuses, organoids, various non-human animals and certain kinds of AI.[9] Similarly, a prominent report led by Long and Sebo argues that although we face significant uncertainty regarding AI consciousness, there is nonetheless an obligation to take AI welfare seriously.[10]

The Precautionary Principle suggests we should take measures to minimise suffering in any AI that has a significant probability of sentience. This means that we must set a probability-threshold for when precautionary measures kick in. If an AI has a probability of sentience that is below that threshold, then no precautionary measures are needed. But if it has a probability above that threshold then proportionate precautions must be taken. The question of where to set this threshold is a very thorny one. Rather than putting a number on it, Birch suggests that what we should look for is a chance of sentience that it would be ‘irresponsible to ignore’.[11]

It is important that the Precautionary Principle only mandates policies that are relatively low-cost. So although we should take measures to protect the welfare of AI that might be conscious, we should not take measures that entail unacceptable costs. This means that if AI turns out to be sentient, its interests are reasonably well-protected. But it also means that if it turns out not to be sentient then the wasted resources will not to be too detrimental to the interests of humans (or other sentient beings). Applying this principle by no means avoids the moral dilemma. After all, if AI is sentient then it could well deserve *more* than just low-cost precautions. And if the AI is not sentient then even low-cost precautions are wasted resources that could have been better spent elsewhere. Nevertheless, the Precautionary Principle promises to *mitigate* the risks entailed by the dilemma.
The Precautionary Principle is intended to show us how to act responsibly in the face of uncertainty regarding moral patienthood. However, when we try to implement the Precautionary Principle we find that it too is compromised by deep uncertainty.

[9] Birch, 2024.
[10] Long *et al.* 2024.
[11] Birch, 2024, 124.

How exactly should we determine the probability of an AI being sentient? Assessments of sentience in animals are already extremely difficult, but assessment of sentience in AI is even harder. Although we are looking here for the *likelihood* of sentience rather than a definitive yes/no answer, such probabilistic assessments are still riddled with uncertainty. Deep problems persist regarding what should count as a marker of consciousness, how those markers should be weighted, and whether those markers are even applicable to AI. Any assessment of the likelihood of an AI being conscious would have a sizeable margin of error. But given the aforementioned issues around thresholds for precaution, such margins of error will be problematic. If we cannot be confident in placing an AI on one side of the probabilistic threshold or the other, then we cannot be confident in our application of the Precautionary Principle.

We might hope that as measures improve, we will be able to make assessments of sentience more confidently. I would argue that such optimism is misplaced.[12] The reason we have such difficulty assessing sentience is that consciousness is deeply resistant to scientific enquiry. These difficulties in measurement are rooted in the deeper 'hard problem' of consciousness, and the hard problem is going nowhere fast.

## 3 THE AVOIDANCE STRATEGY AND ITS DISCONTENTS

In light of the issues above, one might be tempted by a more circumspect approach. Once we appreciate the seriousness of the moral dilemma with which potentially sentient AI presents us, perhaps the most responsible course of action is to avoid being presented with that dilemma in the first place. This is the thought the underwrites Metzinger's proposed *moratorium* on the development of potentially conscious AI.[13] In the same spirit, Schwitzgebel and Garza argue that we should avoid creating AI the consciousness of which is uncertain.[14] I call this the Avoidance Strategy.

Like the Precautionary Principle, the Avoidance Strategy aims to prevent situations in which we cause sentient AI to suffer. The Precautionary Principle does this by saying that when potentially conscious AI is created, we must take appropriate precautions to avoid harming its welfare. The Avoidance Strategy does this by mandating that we do not create potentially conscious AI in the first place. As we saw, a key risk with the Precautionary Principle was that we might end up taking precautions to protect the welfare of AI that is not actually conscious. The Avoidance Strategy requires no such precautions to be taken. Instead, we can prevent AI suffering simply by not creating AI that is potentially conscious. Unlike with issues of animal welfare where the animals in question already exist, when it comes to AI welfare we have the option of avoiding the creation of these morally problematic entities.

The Avoidance Strategy is premised on the idea that there is too much uncertainty around AC for us to be able to deal responsibly with potentially conscious AC. I argue, however, that issues of uncertainty still present a serious problem for the Avoidance Strategy. The rule is that if we are certain that the AI we are developing will not be conscious then we can continue as normal, but if we are uncertain about its consciousness then we should avoid making it. An advantage of this rule is that it never requires us to make confident assessments of artificial consciousness: the rule applies as soon as there is uncertainty about the AI being conscious. The problem is that this requires a boundary to be drawn between cases in which we are certain and cases in which we are not and such boundary is hard to come by.

Because uncertainty around AC runs so deep, we cannot even agree on how uncertain we are. For instance, some parties say that we can *confidently* rule out current LLMs being conscious whereas other parties say that there is uncertainty over this. Put another way, the proposal requires a boundary between certain cases and uncertain cases but that boundary is itself something we are deeply uncertain about. So instead of a recalcitrant disagreement about which AIs are likely to be conscious, this strategy leaves us with recalcitrant disagreement about which AIs are such that we are uncertain of their consciousness.

This problem of meta-uncertainty parallels the problem faced by the Precautionary Principle. The Precautionary Principle tries to acknowledge uncertainty around AC by suggesting probabilistic assessments of AC accompanied by proportionate precautions. But the uncertainty of those probabilistic assessments makes the approach untenable. The Avoidance Strategy tries to acknowledge uncertainty around AC by suggesting a ban on AI the consciousness of which is uncertain. But uncertainty regarding the boundary between certain cases and uncertain cases makes the approach untenable.

[12] McClelland, 2025.
[13] Metzinger, 2021.
[14] Schwitzgebel and Garza, 2020.

## 4 THE SHIFT TO VALENCE

As we have seen, the moral dilemma regarding AC rests on uncertainty around AI sentience. Uncertainty around AI sentience rests on uncertainty around AI consciousness: the main reason we cannot assess AI sentience with any confidence is that we cannot assess AI consciousness with any confidence. If we could somehow assess an AI's prospects of sentience *without* having to assess its prospects of consciousness, we would be in a much better position. On the face of it this is absurd: sentience is a kind of consciousness, so any assessment of sentience must rely on an assessment of sentience. However, further reflection shows that it is far from absurd.

To be sentient an AI must be conscious and have valenced mental states. We have already seen the difficulties with assessing how likely an AI is to be conscious. But we can instead shift to assessing whether the AI has valenced states. Whether an entity has valenced states is something that can be assessed without having to assess if for consciousness. If we know that an entity has valenced states that would not mean that it is sentient. Instead, it means that it *would* be sentient *if* it were conscious. In other words, it has the kind of states that would be positive or negative to be in if consciously experienced. However, if we know that an entity *lacks* valenced states, we can rule out its sentience without having to say anything about its consciousness. Such an entity would either not be conscious or would have a Vulcan consciousness with no valenced experiences. Either way, it would be insentient.

This line of thought might be easier to understand if we step away from valence and AI. Let us instead think about visual experiences in sharks. Do sharks experience colour? This is a question about the conscious states of sharks. As such, it might seem that answering the question would require us to assess how likely the shark is to be conscious. However, experiencing colour requires two things: i) visually representing colour and ii) those visual representations being conscious. If we can rule out sharks satisfying the first requirement, we can safely ignore the second. As it happens, sharks have monochromatic vision. They do not have the cone cells required for colour discrimination so do not visually represent colour. That means we can rule out sharks having colour experiences without having to take a stand on shark consciousness.

Applying this to sentience, it should be possible to assess whether an AI is in the kind of state that *would* be positive or negative to experience *if* it were conscious. After all, what explains a mental state being conscious is different to what explains it having valenced content. We should expect a theory of valence to be able to explain the latter without saying anything about the former. Because these two kinds of explanation come apart, it is possible to assess what kind of content a subject's experience would have without committing to whether that subject is conscious. And this is good news for the study of AI consciousness where it is especially difficult to assess whether the AI is conscious. Indeed, Chrisley makes the case for an AC research program that explores what kinds of conscious experience AI would have (if conscious) while bracketing the vexed question of what explains consciousness.[15]

A complication here is that the very idea of valence might be bound up with consciousness in a way that colour perception is not. With perceptions of colour, we can make sense of such a representational state occurring consciously or non-consciously. But what does it mean for a feeling of joy or of pain to exist unconsciously? If unconscious valence is impossible, then attributions of valenced states would rely on attributions of consciousness after all.

There are two ways round this worry. The first is to defend the idea of unconscious valence. A growing body of empirical work suggests that unconscious states have valences that guide behaviour without the subject ever experiencing that state. Winkielman and Berridge for example, make a strong case for unconscious emotions.[16] This suggests that it is quite possible to assess the valence of a mental state without committing to whether it is conscious.

A critic, however, may insist that such unconscious states would not truly have valence. Solms for instance, rhetorically asks 'How can you have a feeling without feeling it?'[17] This leads us to a second possible response. Instead of appealing to nonconscious valenced states, we can appeal to states that *would* be valenced *if* they were conscious. Conditionalised attributions of valence do not actually require the possibility of nonconscious valence. All that is required is the identification of states that would be valenced if they were conscious. For instance, this nonconscious emotion-like state is one that *would* be positive for the subject *if* conscious but in its unconscious form would just have a kind of proto-valence. Someone committed to valence being essentially conscious has no reason to resist such conditionalisations. For convenience though, I will avoid talking of states-that-would-be-valenced-if-conscious and simply talk of valenced states.

[15] Chrisley, 2008.
[16] Winkielman and Berridge, 2004.
[17] Solms, 2021, 156.

Another possible objection to the strategy of shifting to valence assessments is that assessments of valence are inevitably uncertain as well. Given that the problems established for the Precautionary Principle and Avoidance Strategy pertained to uncertainty, any uncertainty around valence-attribution would itself be problematic.

Here I think it is important to recognise that there are different levels of uncertainty. There will, of course, be uncertainty in assessments of how likely an AI is to have valenced states (as I will discuss in Section 7). Such uncertainty is an inevitable feature of empirical inquiry. The point is that this uncertainty is far less deep than that faced by assessments of artificial consciousness. Efforts to test for consciousness are stymied by the hard problem – the unique resistance of phenomenal consciousness to scientific explanation – and this means we should have little confidence in assessments of artificial consciousness. Efforts to assess whether an AI has valenced states, by contrast, just face the kinds of epistemic problem we find in the course of ordinary empirical enquiry.

Crucially, uncertainty in valence-attributions is the kind of uncertainty that we can expect to decrease through further enquiry. The proposal is to shift our attention from the *intractable* question of artificial consciousness to the *tractable* question of artificial valence. Tractable questions can still be very difficult but are a far wiser target for our research efforts than intractable ones (see Comsa 2026). The objective was never to escape uncertainty. It was just to avoid the deep uncertainty entailed by assessments of consciousness.

Overall, the shift to valence is a promising way of approaching questions of artificial sentience. In the following two sections I consider what difference this shift would make to the Precautionary Principle and Avoidance Strategy respectively.

## 5 THE REVISED PRECAUTIONARY PRINCIPLE

What does the foregoing mean for the application of the Precautionary Principle? The Original Precautionary Principle asks how likely an AI is to be sentient i.e. how likely it is to be a conscious subject with valenced experience. Shifting to valence, a Revised Precautionary Principle would ask how likely an AI is to have states that *would* constitute valenced experiences *if* they were conscious. This avoids the need for assessments of AI consciousness. To unpack this, we should consider what the Revised Precautionary Principle would say in different kinds of case.

First, consider cases where an AI is deemed *unlikely* to have valenced states. Here the principle yields a clear verdict: the AI has a low chance of being sentient, irrespective of how likely it is to be conscious, so no precautions need to be taken. This has a clear advantage over the Original Precautionary Principle: ruling out valence should be easier than ruling out consciousness. The way is open to reach a 'no precaution needed' verdict without having to do any assessment of an AI's likelihood of consciousness.

Second, consider cases where an AI is deemed *likely* to have valenced states conscious. Here our obligations are less clear and some of the difficulties faced by the original Precautionary Principle persist.

One option is to say that if an AI is deemed likely to have valenced states then we should treat it as if it is sentient just in case. This would clearly mitigate the risk of causing undue suffering, but it might come at too high a cost. As with the Original Precautionary Principle, there is a risk of wasting resources on protecting AI that turns out not to be sentient. In fact, a case can be made for this risk being *greater* for the Revised Precautionary Principle. That is because it would encompass any AI with a significant chance of having valenced states even if that AI has a *low* chance of being conscious. Depending on one's theory of valence, this could include a very large number of machines. By no longer including an assessment of consciousness in the criteria, the Revised Precautionary Principle could result in a large number of AIs being covered and so a greater risk of misdirected precautions.

This leads us to a second option, which is to say that if an AI is likely to have valenced states then we should assess how likely it is to be conscious. If the odds of it being conscious are above a certain threshold, then the precautions kick in. And if the odds are below that threshold, the precautions do not kick in. The problem here is that it takes us back to the same old challenge of uncertainties around AI consciousness.

Overall then, the Revised Precautionary Principle is not really an improvement. It has the advantage of being able to rule *out* precautions for AIs that lack valenced states. But when it comes to which AIs to rule *in* the old problems persist.

## 6 THE REVISED AVOIDANCE STRATEGY

What would the shift to valence mean for the Avoidance Strategy? The proposal would be to avoid creating AI with valenced states and thus avoid the moral dilemma of how to treat potentially conscious AI. If an AI is deemed not to have valenced states, then we can carry on without any worries. Such AI would either be non-conscious or would only have a valence-free Vulcan consciousness. Either way, it would not be sentient and so would not be a moral patient. If an AI is deemed to have valenced states, then we should avoid creating AI of that kind.

How does this improve on the Original Avoidance Strategy? The problem with the original version is that it required us to draw a line between AI that is certainly not conscious and AI the consciousness of which is uncertain. The Revised Avoidance Strategy requires no such line to be drawn. Instead, the line is between AI with valenced states and AI without valenced states. As discussed, there will still be a degree of uncertainty about how to draw this line. However, it avoids the much *deeper* uncertainty surrounding artificial consciousness. By shifting from the intractable question of AI consciousness to the tractable question of AI valence, we can avoid the risk of AI sentience more confidently.

Worries might be raised about the potential *costs* of implementing this strategy. One of the problems faced by the Precautionary Principle is that precautions designed to protect potentially conscious AI would be an unacceptable cost if it turned out that the AI was not conscious. Even low-cost precautions could add up to something detrimental to the interests of genuine moral patients. Does a parallel problem apply to the Avoidance Strategy?

Here the costs are a little different. Because this proposal bans the development of potentially sentient AI, no costs are entailed regarding the protection of such entities. However, by blocking the development of such AI there is an *opportunity* cost that must be factored in. What if advanced AI requires an architecture with valenced states i.e. states that would be positive or negative to experience if conscious? Here the Revised Avoidance Strategy would deprive us of the advantages to be gained from advanced AI. Of course, another possibility is that progress in AI is in no way dependent on AI having valenced states. If that's the case, then the proposal comes with no opportunity cost.

This is a difficult empirical question that I cannot hope to answer in this paper. Instead, I will acknowledge the possibility of such opportunity costs but suggest that the burden of proof is on the critic to show why progress in AI would require creating AI with valenced states.

Overall, the Revised Avoidance Strategy is a promising approach with advantages over the Original Avoidance Strategy and both versions of the Precautionary Principle.

## 7 THE PROSPECTS OF AI VALENCE

In the foregoing I have said nothing about how likely AI is to have valenced states or how to test for it. Taking a stand on this is beyond the scope of the paper. Nevertheless, it is worth noting the recent flurry of research in the area.

Sofroniew *et al* explore emotion-like traits in Claude Sonnet 4.5 and argue that the LLM has a host of 'functional emotions' (which the authors are careful to distinguish from subjectively experienced emotions).[18] Another strand of research explores the preferences of LLMs. For example, Keeling *et al* showed how LLMs were able to perform motivational trade-offs between different states that were stipulated to be pleasurable/painful.[19] Ensign, Sleight and Fish use LLM decisions to leave chats as an indicator of preferences and found patterns in the kinds of conversation that the LLM finds aversive.[20] On the more positive side, there is continuing interest in the 'bliss attractor state' first identified in Claude 4's system card.[21] When two instances of Claude are given free reign to talk, their conversations typically show a distinctive pattern: they become increasingly mystical and positive and terminate in what seem to be expressions of meditative bliss.

This research is very provocative and is exactly the kind of thing we should be attending to. But does it really show that the LLMs in question have valanced states? And does it reveal what those valanced states are? Here we face a plethora of methodological challenges.

Some of these are general challenges that we also face when exploring sentience in animals. For instance, how do we avoid anthropomorphism? This is the risk of over-attributing valenced states because of superficial similarities to human expressions of valenced states. Conversely, how do we avoid anthropocentrism? This is the risk of being sensitive only to valenced states that are of a familiar kind and/or expressed in familiar ways and failing to detect valanced states that are of an unfamiliar kind and/or expressed in unfamiliar ways. Moreover, there are ethical issue regarding how to test for negative valence without bringing about negatively valenced states and thereby causing undue suffering.

---

[18] Sofroniew *et al,* 2026.
[19] Keeling *et al*, 2024.
[20] Ensign, Sleight and Fish, 2025.
[21] Anthropic, 2025.

Some of the challenges above could well be more acute in an AI context. There are also some further challenges more distinctive to AI. For example, how do we rule out the possibility that AI lacks valenced states but has learned from its training data to respond to prompts in valenced-like ways? Relatedly, how do we rule out the possibility that the LLM is engaging in a kind of 'roleplay' where it adopts the persona of an entity with particular preferences while actually having different preferences or even no preferences at all?

AI also raises deeper questions regarding the very nature of valenced states. For example, embodiment is thought to play a central role in the valenced states of organisms. Positive feelings of joy and negative feelings of anger each have a key bodily component. What, then, should we make of valence in disembodied AIs? Do we have to expand our outlook to encompass valenced states without an embodied component? Or should we say that lacking a body precludes such AIs from having valenced states? And when it comes to embodied AI, what kind of embodiment would be relevant to attributions of valence?

These are challenging questions indeed. But if we are concerned with AI welfare, we will be better off focusing our resources on these questions rather than the intractable question of artificial consciousness. Chipping away at difficult problems is arduous but worthwhile. In contrast, repeatedly butting into the hard problem is futile. The shift to valence is thus a shift in the right direction.

## 8 CONCLUSION

The problem faced by any attempt to deal responsibly with the risk of sentient AI is that of uncertainty around AI consciousness. The original Precautionary Principle and Avoidance Strategy were designed to accommodate that uncertainty, but each ended up facing issues of uncertainty of their own. By shifting from questions of consciousness to questions of valence, we can make assessments of AI sentience more tractable. Revising the Precautionary Principle in a way that avoids questions of consciousness turned out not to be feasible. But revising the Avoidance Strategy in that way emerged as a very promising option. This not only gives us a better way of dealing responsibly with the risk of AI sentience but points us towards a more promising research program. The proposed path ahead is still a difficult one, but I hope to have shown that it has considerable advantages over the path we are currently on.